\documentclass[10pt,twocolumn,letterpaper]{article}

\usepackage{cvpr}
\usepackage{times}
\usepackage{epsfig}
\usepackage{graphicx}
\usepackage{amsmath}
\usepackage{amssymb}
\usepackage{ulem}
\usepackage{multirow}
\usepackage{multicol}
\usepackage{ulem} 
\usepackage{authblk}

\usepackage[breaklinks=true,bookmarks=false]{hyperref}

\cvprfinalcopy 

\def\cvprPaperID{189} 

\ifcvprfinal\fi
\begin{document}

\newcommand{\etals} {\textit{et al.}}

\title{High-Order Information Matters: Learning Relation and Topology \\ for Occluded Person Re-Identification}

\author[1\thanks{Equal contribution, works done as interns in Megvii Research.}]{Guan'an Wang}
\author[3$^*$]{Shuo Yang}
\author[2]{Huanyu Liu}
\author[2]{Zhicheng Wang}
\author[1]{Yang Yang}
\author[3]{Shuliang Wang}
\author[2]{Gang Yu}
\author[2]{Erjin Zhou}
\author[2]{Jian Sun}
\affil[  ]{\textsuperscript{1}Institute of Automation, CAS \quad
\textsuperscript{2}MEGVII Technology \quad \textsuperscript{3}Beijing Institute of Technology
}
\affil[ ]{\tt\small
 \textsuperscript{1}wangguanan2015@ia.ac.cn
 \textsuperscript{2}\{liuhuanyu,wangzhicheng,yugang,zej,sunjian\}@megvii.com \textsuperscript{3}\{shuoyang,slwang2011\}@bit.edu.cn
 \textsuperscript{1}\{yang.yang\}@nlpr.ia.ac.cn
}
\renewcommand\Authands{ and }

\maketitle

\begin{abstract}
Occluded person re-identification (ReID) aims to match occluded person images to holistic ones across dis-joint cameras. In this paper, we propose a novel framework by learning high-order relation and topology information for discriminative features and robust alignment.
%
%
At first, we use a CNN backbone and a key-points estimation model to extract semantic local features. Even so, occluded images still suffer from occlusion and outliers.
Then, we view the local features of an image as nodes of a graph and propose an adaptive direction graph convolutional (ADGC) layer to pass relation information between nodes. The proposed ADGC layer can automatically suppress the message passing of meaningless features by dynamically learning direction and degree of linkage.
When aligning two groups of local features from two images, we view it as a graph matching problem and propose a cross-graph embedded-alignment (CGEA) layer to jointly learn and embed topology information to local features, and straightly predict similarity score. The proposed CGEA layer not only take full use of alignment learned by graph matching but also replace sensitive one-to-one matching with a robust soft one.
Finally, extensive experiments on occluded, partial, and holistic ReID tasks show the effectiveness of our proposed method. 
Specifically, our framework significantly outperforms state-of-the-art by $6.5\%$ mAP scores on Occluded-Duke dataset.
Code is available at \url{https://github.com/wangguanan/HOReID}.

\end{abstract}

\section{Introduction}

\begin{figure}[t]
\includegraphics[width=\linewidth]{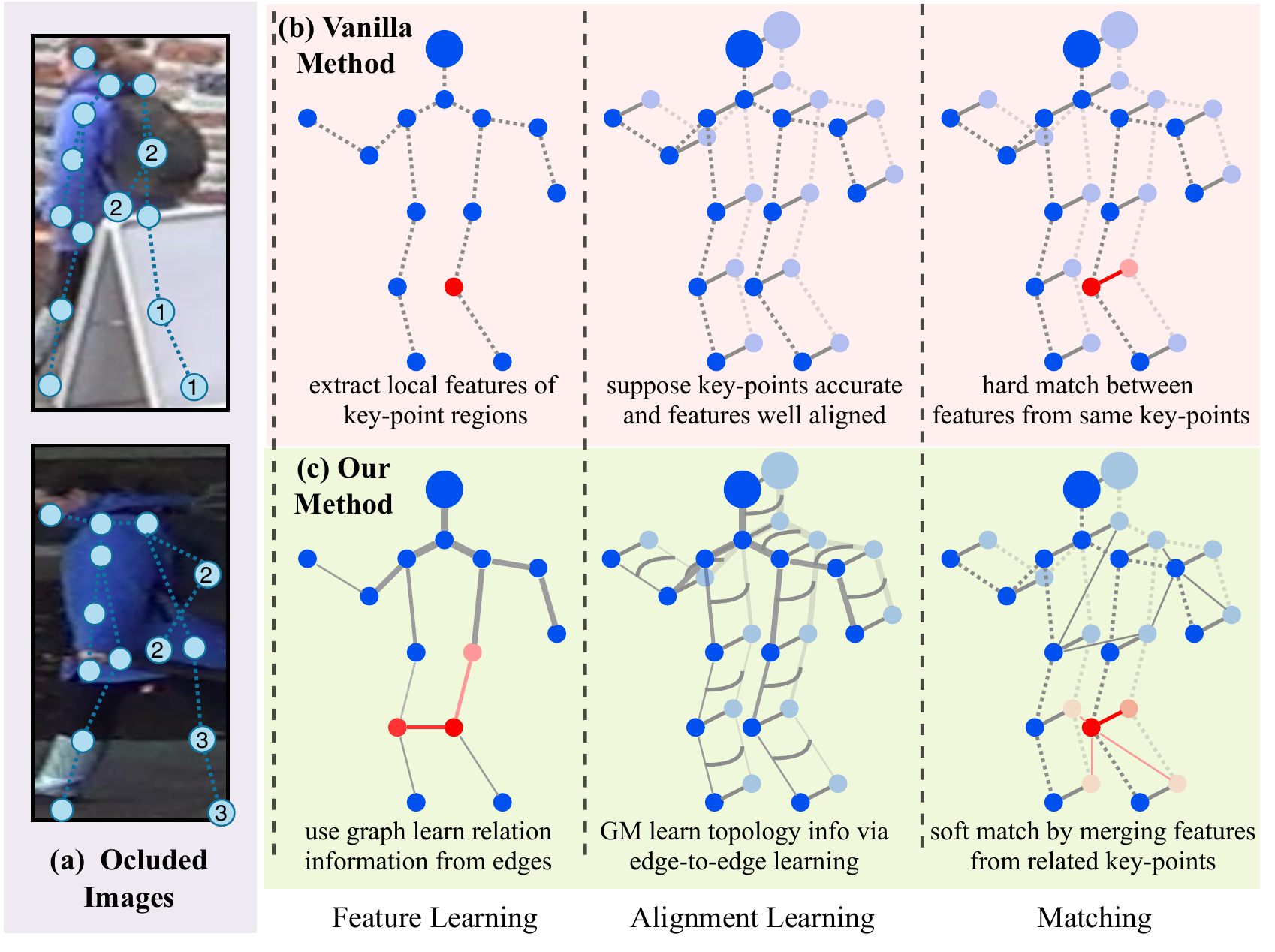}
\caption{
Illustration of high-order relation and topology information.
(a) In occluded ReID, key-points suffer from occlusions (\textcircled{1}\textcircled{2}) and outliers (\textcircled{3}).
(b) Vanilla method relies on one-order key-points information in all three stages, which is not robust.
(c) Our method learn features via an graph to model relation information , and view alignment as a graph matching problem to model topology information by learning both node-to-node and edge-to-edge correspondence.
}
\label{fig:intro}
\end{figure}

Person re-identification (ReID) \cite{gong2014person,zheng2016person} aims to match images of a person across dis-joint cameras, which is widely used in video surveillance, security and smart city.
Recently, various of methods \cite{ma2014covariance,yang2014salient,liao2015person,zheng2013reidentification,koestinger2012large,liao2015efficient,zheng2016person,hermans2017defense,sun2018beyond} have been proposed for person ReID.
However, most of them focus on holistic images, while neglecting occluded ones, which may be more practical and challenging.
As shown in Figure \ref{fig:intro}(a), persons can be easily occluded by some obstacles (\textit{e.g.} baggage, counters, crowded public, cars, trees) or walk out of the camera fields, leading to occluded images. 
Thus, it is necessary to match persons with occluded observation, which is known as occluded person Re-ID problem \cite{zhuo2018occluded,miao2019PGFA}.

Compared with matching persons with holistic images, occluded ReID is more challenging due to the following reasons \cite{zheng2015partial, zhuo2018occluded}:
(1) 
With occluded regions, the image contains less discriminative information and is more likely to match wrong persons.
(2)
Part-based features have been proved to be efficient~\cite{sun2018beyond} via part-to-part matching. But they require strict person alignment in advance, thus cannot work very well in seriously occluded situations. 
Recently, many occluded/partial person ReID methods \cite{zhuo2018occluded,zhuo2019novel,miao2019PGFA,he2019foreground-aware, he2018deep, sun2019perceive, luo2019stnreid} are proposed, most of them only consider one-order information for feature learning and alignment. 
%
For example, the pre-defined regions~\cite{sun2018beyond}, poses~\cite{miao2019PGFA} or human parsing~\cite{he2019foreground-aware} are used to for feature learning and alignment. 
We argue that besides one-order information, high-order one should be imported and may work better for occluded ReID. 

In Figure \ref{fig:intro}(a), we can see that  key-points information suffers from occlusion (\textcircled{1}\textcircled{2}) and outliers (\textcircled{3}).
For example, key-points \textcircled{1} and \textcircled{2} are occluded, leading to meaningless features. 
Key-points \textcircled{3} are outliers, leading to misalignment.
A common solution is shown in Figure \ref{fig:intro}(b). It extracts local features of key-point regions, supposes all key-points are accurate and local features well aligned. In this solution, all three stages rely on the one-order key-points information, which is not very robust.
In this paper, as shown in Figure \ref{fig:intro}(c), we propose a novel framework for both discriminative feature and robust alignment.
In feature learning stage, we view local features of an image as nodes of a graph to learn relation information. By passing message in the graph, the meaningless features caused by occluded key-points can be improved by their neighbor meaningful features.
In alignment stage, we use graph matching algorithm \cite{zanfir2018deep} to learn robust alignment. Besides aligning with node-to-node correspondence, it models extra edge-to-edge correspondence.
We then embed the alignment information to features by constructing a cross-images graph, where node message of an image can be passed to nodes of the other images. Thus, the features of outlier key-points can be repaired by its corresponding features on the other image.
Finally, instead of computing similarity with predefined distance, we use a network to learn similarity supervised by a verification loss.

\begin{figure*}[t]
\centering
\includegraphics[width=\textwidth]{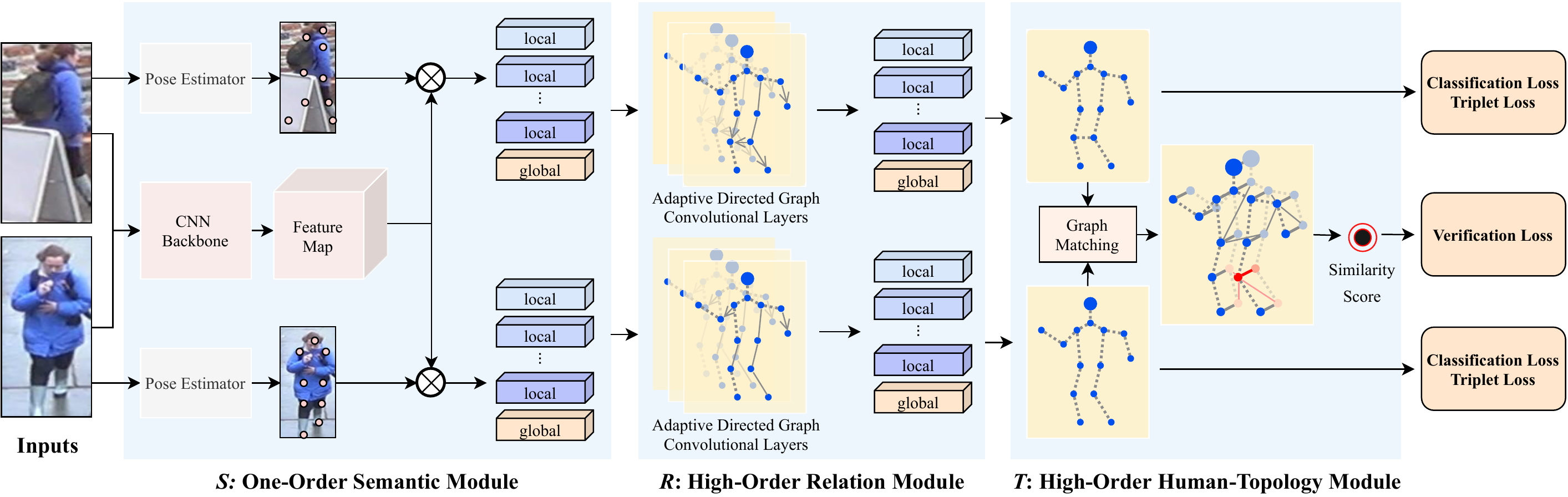}
\caption{Illustration of our proposed framework. It consists of an one-order semantic module $\mathcal{S}$, a high-order relation module $\mathcal{R}$ and a high-order topology module $\mathcal{T}$. The module $\mathcal{S}$ learns semantic local features of key-point regions. In $\mathcal{R}$, we view the local features of an image as nodes of a graph and propose an adaptive direction graph convolutional (ADGC) layer to pass relation information between nodes. In $\mathcal{T}$, we view alignment problem as a graph matching problem and propose a cross-graph embedded-alignment (CGEA) layer to joint learn and embed topology information to local features, and straightly predict similarity scores.}
\label{fig:pipeline}
\end{figure*}

Specifically, we propose a novel framework jointly modeling high-order relation and human-topology information for occluded person re-identification.
As shown in Figure \ref{fig:pipeline}, our framework includes three modules, \textit{i.e.} one-order semantic module ($\mathcal{S}$), high-order relation module ($\mathcal{R}$) and high-order human-topology module ($\mathcal{T}$).
\textbf{(1)}
In the $\mathcal{S}$, we utilize a CNN backbone to learn feature maps and a human key-points estimation model to learn key-points. Then we can extract semantic features of corresponding key-points.
\textbf{(2)}
In the $\mathcal{R}$, we view the learned semantic features of an image as nodes of a graph and propose an adaptive-direction graph convolutional (ADGC) layer to learn and pass messages of edge features. The ADGC layer can automatically decide the direction and degree of every edge. Thus it can promote the message passing of semantic features and suppress that of meaningless and noisy ones. At last, the learned nodes contain both semantic and related information.
\textbf{(3)}
In the $\mathcal{T}$, We propose a cross-graph embedded-alignment (CGEA) layer. It takes two graphs as inputs, learns correspondence of nodes across the two graphs using graph-matching strategy, and passes messages by viewing the learned correspondence as an adjacency matrix. Thus, the related features can be enhanced, and alignment information can be embedded in features. Finally, to avoid hard one-to-one alignment, we predict the similarity of the two graphs by mapping them to a logit and supervise with a verification loss.

The main contributions of this paper are summarized as follows:
\textbf{(1)}
A novel framework of jointly modeling high-order relation and human-topology information is proposed to learn well and robustly aligned features for occluded ReID. To our best of our knowledge, this is the first work that introduces such high-order information to occluded ReID.
\textbf{(2)}
An adaptive directed graph convolutional (ADGC) layer is proposed to dynamically learn the directed linkage of the graph, which can promote message passing of semantic regions and suppress that of meaningless regions such as occlusion or outliers. With it, we can better model the relation information for occluded ReID.
\textbf{(3)}
A cross-graph embedded-alignment (CGEA) layer conjugated with verification loss is proposed to learn feature alignment and predict similarity score. They can avoid sensitive hard one-to-one person matching and perform a robust soft one.
\textbf{(4)}
Extensive experimental results on occluded, partial, and holistic ReID datasets demonstrate that the proposed model performs favorably against state-of-the-art methods. Especially on the occluded-Duke dataset, our method significantly outperforms state-of-the-art by at least 3.7\% and 6.5\% in terms of Rank-1 and mAP scores.

\section{Related Works}


\textbf{Person Re-Identification.}
Person re-identification addresses the problem of matching pedestrian images across disjoint cameras \cite{gong2014person}. 
The key challenges lie in the large intra-class and small inter-class variation caused by different views, poses, illuminations, and occlusions.
Existing methods can be grouped into hand-crafted descriptors \cite{ma2014covariance,yang2014salient,liao2015person}, metric learning methods \cite{zheng2013reidentification,koestinger2012large,liao2015efficient} and deep learning algorithms \cite{zheng2016person,hermans2017defense,sun2018beyond,Wang_2019_ICCV,Wang_2020_AAAI,lu2020crossmodality}. 
%
%
%
All those ReID methods focus on matching holistic person images, but cannot perform well for the occluded images, which limits the applicability in practical surveillance scenarios.
%
%
%
%
%

\textbf{Occluded Person Re-identification.}
Given occluded probe images, occluded person re-identification \cite{zhuo2018occluded} aims to find the same person of full-body appearance in dis-joint cameras. This task is more challenging due to incomplete information and spatial misalignment.
Zhuo \etals \cite{zhuo2018occluded} use occluded/non-occluded binary classification(OBC) loss to distinguish the occluded images from holistic ones. In their following works, a saliency map is predicted to highlight the discriminative parts, and a teacher-student learning scheme further improves the learned features. 
Miao \etals \cite{miao2019PGFA} propose a pose guided feature alignment method to match the local patches of probe and gallery images based on the human semantic key-points. And they use a pre-defined threshold of key-points confidence to determine whether the part is occluded or not. 
Fan \etals \cite{fan2018scpnet} use a spatial-channel parallelism network (SCPNet) to encode part features to specific channels and fuse the holistic and part features to get discriminative features. 
Luo \etals \cite{luo2019stnreid} use a spatial transform module to transform the holistic image to align with the partial ones, then calculate the distance of the aligned pairs.
Besides, several efforts are put on the spatial alignment of the partial Re-ID tasks. 

\textbf{Partial Person Re-Identification.}
Accompanied by occluded images, partial ones often occur due to imperfect detection and outliers of camera views. Like occluded person ReID, partial person ReID \cite{zheng2015partial} aims to match partial probe images to gallery holistic images. 
Zheng \etals \cite{zheng2015partial} propose a global-to-local matching model to capture the spatial layout information. 
He \etals \cite{he2016deep} reconstruct the feature map of a partial query from the holistic pedestrian, and further improve it by a foreground-background mask to avoid the influence of backgrounds clutter in \cite{he2019foreground-aware}.
Sun \etals propose a Visibility-aware Part Model(VPM) in \cite{sun2019perceive}, which learns to perceive the visibility of regions through self-supervision. 

Different from existing occluded and partial ReID methods which only use one-order information for feature learning and alignment, we use high-order relation and human-topology information for feature learning and alignment, thus achieve better performance.

\section{The Proposed Method}

This section introduces our proposed framework, including a one-order semantic module ($\mathcal{S}$) to extract semantic features of human key-point regions, a high-order relation module ($\mathcal{R}$) to model the relation-information among different semantic local features, and a high-order human-topology module ($\mathcal{T}$) to learn robust alignment and predict similarities between two images.
The three modules are jointly trained in an end-to-end way.
An overview of the proposed method is shown in Figure \ref{fig:pipeline}.

\textbf{Semantic Features Extraction.}
The goal of this module is to extract one-order semantic features of key-point regions, which is inspired by two cues. Firstly, part-based features have been shown to be efficient for person ReID \cite{sun2018beyond}. Secondly, accurate alignment of local features is necessary in occluded/partial ReID \cite{he2018deep,sun2019perceive,he2019foreground-aware}. Following the ideas above and inspired by recent developments on person ReID \cite{zheng2016person,sun2018beyond,luo2019bag,fu2019horizontal} and human key-points prediction \cite{cao2018openpose, SunXLW19}, we utilize a CNN backbone to extract local features of different key-points.
Please note that although the human key-points prediction have achieved high accuracy, they still suffer from unsatisfying performance under occluded/partial images \cite{li2018crowdpose}.
Those factors lead to inaccurate key-points positions and their confidence. Thus, the following relation and human-topology information are needed and will be discussed in the next section.

Specifically, given a pedestrian image $x$, we can get its feature map $m_{cnn}$ and key-points heat map $m_{kp}$ through the CNN model and key-points model.
Through an outer product ($\otimes$) and a global average pooling operations ($g(\cdot)$), we can get a group of semantic local features of key-points regions $V^{S}_l$ and a global feature $V^{S}_g$. 
The procedures can be formulated in Eq.(\ref{eq:local-global-features}), where $K$ is key-point number, $v_k \in R^c$ and $c$ is channel number.
Note that $m_{kp}$ is obtained by normalizing original key-points heatmap with a softmax function for preventing from noise and outliers. This simple operation is shown to be effective in experimet section.
\begin{equation}
\begin{aligned}
V^{S}_l & = \{v_k^{S}\}_{k=1}^{K} = g(m_{cnn} \otimes m_{kp}) \\
V^{S}_g & = {v_{K+1}^{S}} = g(m_{cnn})
\end{aligned}
\label{eq:local-global-features}
\end{equation}

\textbf{Training Loss.}
Following \cite{zheng2016person,hermans2017defense}, we utilize classification and triplet losses as our targets as in Eq.(\ref{eq:semantic-loss}).
Here, $\beta_k=max(m_{kp}[k]) \in [0,1]$ is the $k^{th}$ key-point confidence, and $\beta_{K+1}=1$ for global features, $p_{v^{S}_k}$ is the probability of feature $v^{s}_k$ belonging to its ground truth identity predicted by a classifier, $\alpha$ is a margin, $d_{v_{ak}^s, v_{pk}^s}$ is the distance between a positive pair ($v_{ak}^S$, $v_{pk}^S$) from the same identity, ($v_{ak}^S$, $v_{pk}^S$) is from different identities. 
The classifiers for different local features are not shared.
\begin{small}
\begin{equation}
\begin{aligned}
    \mathcal{L}_{S} & = \frac{1}{K+1} \sum\limits_{k=1}^{K+1} \beta_{k} [ \mathcal{L}_{cls}(v_k^{s}) + \mathcal{L}_{tri}(v_k^{s})] \\
    & = \frac{1}{K+1} \sum\limits_{k=1}^{K+1} \beta_{k} [  -log  p_{v_k^{s}}
    +  |\alpha + d_{v_{ak}^{S}, v_{pk}^S} - d_{v_{ak}^S, v_{nk}^{S}}|_{+}]
\end{aligned}
\label{eq:semantic-loss}
\end{equation}
\end{small}

\subsection{High-Order Relation Learning}

Although we have the one-order semantic information of different key-point regions, occluded ReID is more challenging due to incomplete pedestrian images. Thus, it is necessary to exploit more discriminative features.
We turn to the graph convolutional network (GCN) methods \cite{battaglia2018relational} and try to model the high-order relation information. In the GCN, semantic features of different key-point regions are viewed as nodes. By passing messages among nodes, not only the one-order semantic information (node features) but also the high-order relation information (edge features) can be jointly considered.

However, there is still a challenge for occluded ReID. Features of occluded regions are often meaningless even noisy. When passing those features in a graph, it brings in more noise and has side effects on occluded ReID. 
Hence, we propose a novel adaptive-direction graph convolutional (ADGC) layer to learn the direction and degree of message passing dynamically. With it, we can automatically suppress the message passing of meaningless features and promote that of semantic features. 

\begin{figure}
    \centering
    \includegraphics[scale=0.55]{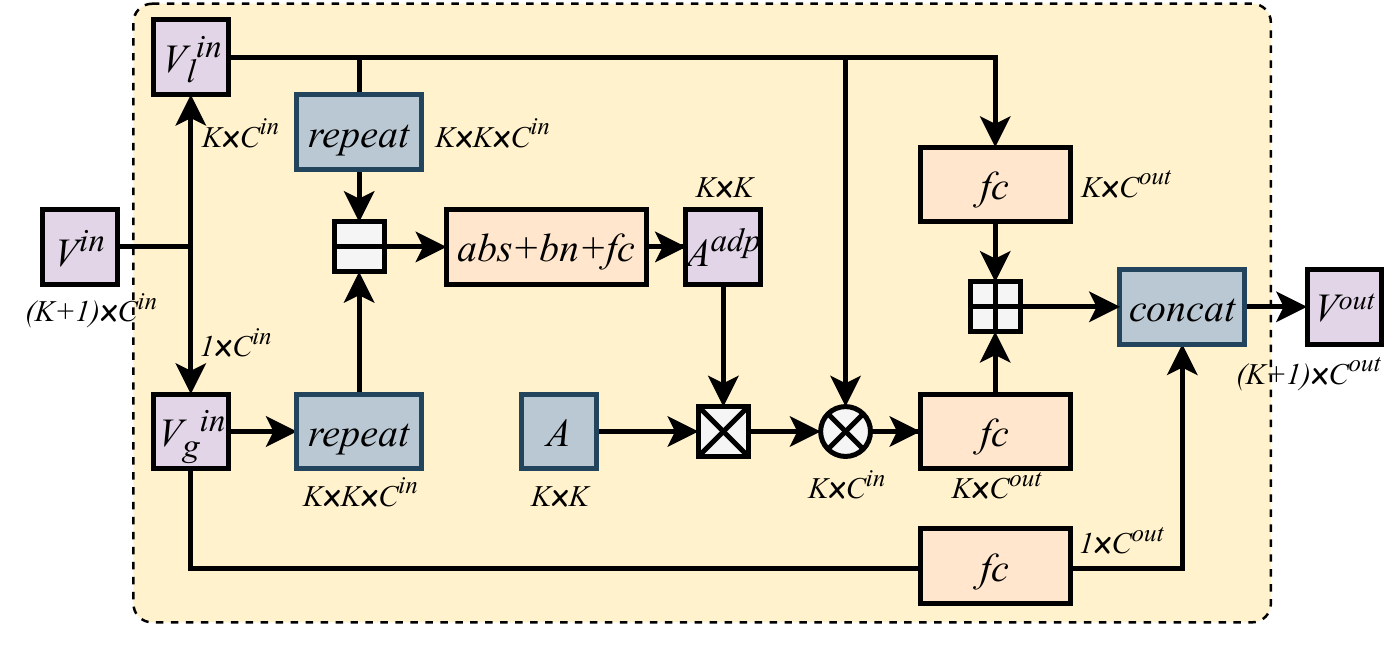}
    \caption{Illustration of the proposed adaptive directed graph convolutional (ADGC) layers. $A$ is a pre-defined adjacent matrix $\boxminus$, $\boxplus$, $\boxtimes$ are element-wise subtraction, add and multiplication. $abs, bn$ and $fc$ are absolution, batch normalization and fully connected layer, $trans$ is transpose. Please refer text for more details.}
    \label{fig:gcn_details}
\end{figure}

\textbf{Adaptive Directed Graph Convolutional Layer.}
A simple graph convolutional layer \cite{kipf2016semi} has two input, an adjacent matrix A of the graph and the features X of all node, output can be calculated by: 
$$O = \hat{A}XW$$
where $\hat{A}$ is normalized version of A and W refers to parameters.

We improve the simple graph convolutional layer by adaptively learning the adjacent matrix (the linkage of node) based on the input features. 
We assume that given two local features, the meaningful one is more similar to the global feature than that of meaningless one.
Therefore, we propose an adaptive directed graph convolutional (ADGC) layer, whose inputs are a global feature $V_g$ and K local features $V_l$, and a pre-defined graph (adjacent matrix is A). We use differences between local features $V_l$ and global feature $V_g$ to dynamically update the edges' weights of all nodes in the graph, resulting $A^{adp}$. 
Then a simple graph convolutional can be formulated by multiplication between  $V_l$ and $A^{adp}$. 
To stabilize training, we fuse the input local features $V_l$ to the output of our ADGC layer as in the ResNet~\cite{he2016deep}. 
Details are shown in Figure \ref{fig:gcn_details}.
Our adaptive directed graph convolutional (ADGC) layer can be formulated in Eq.(\ref{eq:adgc}), where $f_1$ and $f_2$ are two unshared fully-connected layers.
\begin{equation}
    V^{out} = [f_1(A^{adp} \otimes V_l^{in}) + f_2(V_l^{in}), V_g^{in}]
    \label{eq:adgc}
\end{equation}

Finally, we implement our high-order relation module $f_{R}$ as cascade of ADGC layers. Thus, given an image $x$, we can get its semantic features $V^S = \{v_{k}^{S}\}_{k=1}^{K+1}$ via Eq.(\ref{eq:local-global-features}).
Then its relation features $V^R = \{v_{k}^R\}_{k=1}^{K+1}$ can be formulated as below :
\begin{equation}
    V^R = f_{R}(V^S)
    \label{eq:relation-features}
\end{equation}

\textbf{Loss and Similarity.}
We use classification and triplet losses as our targets as in Eq.(\ref{eq:relation-loss}), where the definition of $\mathcal{L}_{ce}(\cdot)$ and $\mathcal{L}_{tri}(\cdot)$ can be found in in Eq.(\ref{eq:semantic-loss}). Note that $\beta_{k}$ is the $k^{th}$ key-point confidence.
\begin{equation}
\begin{aligned}
    \mathcal{L}^{R} & = \frac{1}{K+1} \sum\limits_{k=1}^{K+1} \beta_{k} [ \mathcal{L}_{cls}(v_k^{R}) + \mathcal{L}_{tri}(v_k^{R})] \\
\end{aligned}
\label{eq:relation-loss}
\end{equation}

Given two images $x_1$ and $x_2$, we can get their relation features $V^R_1 = \{v_{1k}^R\}_{k=1}^{K+1}$ and $V^R_1 = \{v_{2k}^R\}_{k=1}^{K+1}$ via Eq.(\ref{eq:relation-features}), and calculate their similarity with cosine distance as in Eq.(\ref{eq:cosine-similarity}).
\begin{equation}
s^{R}_{x_1, x_2} = \frac{1}{K+1} \sum \limits_{k=1} ^{K+1} \
\sqrt{\beta_{1k}\beta_{2k}} \ cosine(v^{R}_{1k},v^{R}_{2k}) 
\label{eq:cosine-similarity}
\end{equation}

\subsection{High-Order Human-Topology Learning}
Part-based features have been proved to be very efficient for person ReID \cite{sun2018beyond,sun2019perceive}.
One simple alignment strategy is straightly matching features of the same key-points. However, this one-order alignment strategy cannot deal with some bad cases such as outliers, especially in heavily occluded cases \cite{li2018crowdpose}.
Graph matching \cite{zanfir2018deep,wang2019learning} can naturally take the high-order human-topology information into consideration. But it can only learn one-to-one correspondence. This hard alignment is still sensitive to outliers and has a side effect on performance.
In this module, we propose a novel cross-graph embedded-alignment layer, which can not only make full use of human-topology information learned by graph matching algorithm, but also avoid sensitive one-to-one alignment.

\textbf{Revision of Graph Matching.}
Given two graphs $G_1=(V_1, E_1)$ and $G_2=(V_2, E_2)$ from image $x_1$ and $x_2$, the goal of graph matching is to learn a matching matrix $U \in [0,1]^{K \times K}$ between $V_1$ and $V_2$.
Let $U \in [0, 1]$ be an indicator vector such that $U_{ia}$ is the matching degree between $v_{1i}$ and $v_{2a}$. A square symmetric positive matrix $M \in R^{KK \times KK}$ is built such that $M_{ia;jb}$ measures how well every pair $(i,j) \in E_1$ matches with $(a,b) \in E_2$. For pairs that do not form edges, their corresponding entries in the matrix are set to 0. The diagonal entries contain node-to-node scores, whereas the off-diagonal entries contain edge-to-edge scores. Thus, the optimal matching $u^*$ can be formulated as below: 
\begin{equation}
    U^* =  \mathop{argmax} \limits_{U} U^T M U,  \ \textit{s.t.} \ ||U|| = 1
    \label{eq:gm}
\end{equation}
Following~\cite{zanfir2018deep}, we parameter matrix $M$ in terms of unary and pair-wise point features. The optimization procedure is formulated by a power iteration and a bi-stochastic operations. Thus, we can optimize $U$ in our deep-learning framework with stochastic gradient descent.
Restricted by pages, we don not show more details of graph matching, please refer to the paper \cite{wang2019learning,zanfir2018deep}.

\begin{figure}
    \centering
    \includegraphics[scale=0.5]{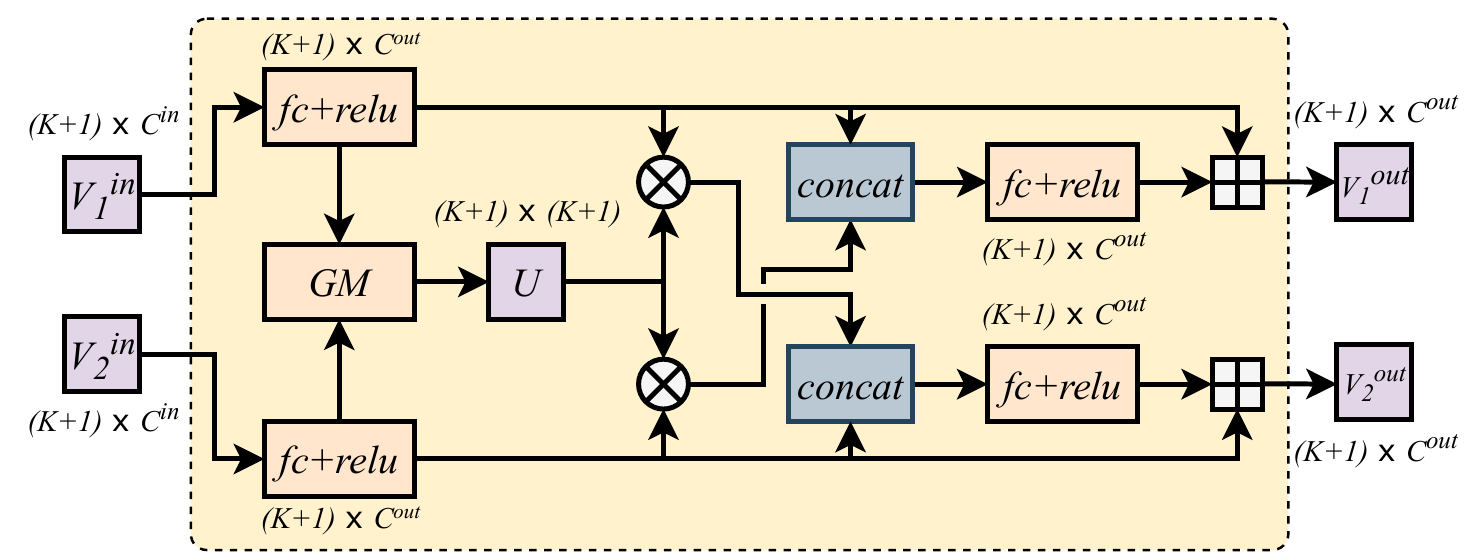}
    \caption{Illustration of the cross-graph embedded-alignment layer. Here, $\otimes$ is matrix multiplication, $fc+relu$ means fully-connected layer and Rectified Linear Unit, $GM$ means graph matching operation, $U$ is the learned affinity matrix. Please refer text for more details.}
    \label{fig:gm_details}
\end{figure}

\textbf{Cross-Graph Embedded-Alignment Layer with Similarity Prediction.}
We propose a novel cross-graph embedded-alignment layer (CGEA) that both considering the high-order human-topology information learned by GM and avoiding the sensitive one-to-one alignment.
The proposed CGEA layer takes two sub-graphs from two images as inputs and outputs the embedded features, including both semantic features and the human-topology guided aligned features.

%
%
%
The structure of our proposed CGEA layer is shown in Figure \ref{fig:gm_details}.  
It takes two groups of features and outputs two groups of features.
Firstly, with two groups of nodes $V_1^{in} \in R^{(K+1) \times C^{in}}$ and $V_2^{in} \in R^{(K+1) \times C^{in}}$, we embed them to a hidden space with a fully-connected layer and a ReLU layer, getting two groups of hidden features $V_1^{h} \in R^{(K+1) \times C^{out}}$ and $V_2^{h} \in R^{(K+1) \times C^{out}}$.
Secondly, we perform graph matching between $V_1^{h}$ and $V_2^{h}$ via Eq.(\ref{eq:gm}), and get an affinity matrix $U^{k\times k}$ between $V_1^{h}$ and $V_2^{h}$. Here, $U(i,j)$ means correspondence between $v_{1i}^{h}$ and $v_{2j}^{h}$. 
Finally, the output can be formulated in Eq.(\ref{eq:CGEA-layer}), where $[\cdot,\cdot]$ means concatenation operation along channel dimension, $f$ is a fully-connected layer.
\begin{equation}
\begin{aligned}
    V^{out}_1 & = f([V_1^{h}, U \otimes V_2^{h}]) + V_1^h  \\
    V^{out}_2 & = f([V_2^{h}, U^T \otimes V_1^{h}]) + V_2^h
\end{aligned}
\label{eq:CGEA-layer}
\end{equation}
We implement our high-order topology module ($\mathcal{T}$) with a cascade of CGEA layers $f_{T}$ and a similarity prediction layer $f_{P}$.
Given a pair of images $(x_1, x_2)$, we can get their relation features $(V^{R}_1, V^{R}_2)$ via Eq.(\ref{eq:relation-features}), and then their topology features of $(V_1^{T}, V_2^{T})$ via Eq.(\ref{eq:topology-features}).
After getting the topology features pair $(V_1^{T}, V_2^{T})$, we can compute their similarity using Eq.(\ref{eq:similarity-prediction}), where $|\cdot|$ is element-wise absolution operation, $f_s$ is a fully-connected layer from $C_{T}$ to $1$, $\sigma$ is sigmoid activation function.
\begin{equation}
    (V_1^T, V_2^T) = F_T(V_1^R, V_2^R)
    \label{eq:topology-features}
\end{equation}
\begin{equation}
    s^{T}_{x_1, x_2} = \sigma(f_s(-|V_1^{T} - V_2^{T}|))
    \label{eq:similarity-prediction}
\end{equation}

\textbf{Verification Loss.}
The loss of our high-order human-topology module can be formulated in Eq.(\ref{eq:loss-topology}), where $y$ is their ground truth, $y=1$ if $(x_1, x_2)$ from the same person, otherwise $y=0$.
\begin{equation}
    \mathcal{L}_{T} = y log s^{T}_{x_1, x_2} + (1-y) log (1-s^{T}_{x_1, x_2})
    \label{eq:loss-topology}
\end{equation}

\section{Train and Inference}

During the training stage, the overall objective function of our framework is formulated in Eq.(\ref{eq:overall-loss}), where $\lambda_{*}$ are weights of corresponding terms. We train our framework end-to-end by minimizing the $\mathcal{L}$.
\begin{equation}
    \mathcal{L} = \mathcal{L}_{S} + \lambda_{R} \mathcal{L}_{R} + \lambda_{T} \mathcal{L}_{T}
    \label{eq:overall-loss}
\end{equation}
For the similarity, given a pair of images $(x_1, x_2)$, we can get their relation information based similarity $s_{x_1, x_2}^{R}$ from Eq.(\ref{eq:cosine-similarity}) and topology information based similarity $s_{x_1, x_2}^{T}$ from Eq.(\ref{eq:similarity-prediction}). The final similarity can be calculated by combing the two kind of similarities.
\begin{equation}
    s = \gamma s_{x_1, x_2}^{R} + (1-\gamma) s_{x_1, x_2}^{T}
    \label{eq:final-similarity}
\end{equation}
When inferring, given an query image $x_q$, we first compute its similarity $x^R$ with all gallery images and get its top $n$ nearest neighbors. Then we compute the final similarity $s_{}$ in Eq.(\ref{eq:final-similarity}) to refine the top $n$.

\section{Experiments}

\subsection{Implementation Details}

\textbf{Model Architectures.}
For CNN backbone, as in~\cite{zheng2016person}, we utilize ResNet50~\cite{he2016deep} as our CNN backbone by removing its global average pooling (GAP) layer and fully connected layer. 
%
%
For classifiers, following~\cite{luo2019bag}, we use a batch normalization layer \cite{ioffe2015batch} and a fully connect layer followed by a softmax function. 
For the human key-points model, we use HR-Net~\cite{SunXLW19} pre-trained on the COCO dataset~\cite{lin2014microsoft}, a state-of-the-art key-points model. The model predicts 17 key-points, and we fuse all key-points on head region and get final $K=14$ key-points, including head, shoulders, elbows, wrists, hips, knees, and ankles.

\textbf{Training Details.}
We implement our framework with Pytorch. 
The images are resized to $256 \times 128$ and augmented with random horizontal flipping, padding 10 pixels, random cropping, and random erasing~\cite{zhong2017random}. When test on occluded/partial datasets, we use extra color jitter augmentation to avoid domain variance. 
The batch size is set to 64 with 4 images per person. 
During the training stage, all three modules are jointly trained in an end-to-end way for 120 epochs with the initialized learning rate 3.5e-4 and decaying to its 0.1 at 30 and 70 epochs. 
Please refer our code\footnote{\url{https://github.com/wangguanan/HOReID}} for implementation details.

\textbf{Evaluation Metrics}. 
We use standard metrics as in most person ReID literatures, namely Cumulative Matching Characteristic (CMC) curves and mean average precision (mAP),  to evaluate the quality of different person re-identification models. All the experiments are performed in single query setting.

\begin{table}[t]
\begin{center}
\small
\label{tab:datasets}
\scalebox{0.97}{
\begin{tabular}{c|c|c|c}
\hline
\hline
\multicolumn{1}{c|}{\multirow{2}{*}{Dataset}} & \multicolumn{1}{c|}{\multirow{2}{*}{\begin{tabular}[c]{@{}c@{}}Train Nums\\ (ID/Image)\end{tabular}}} & \multicolumn{2}{c}{Testing Nums (ID/Image)}                \\ \cline{3-4} 
\multicolumn{1}{c|}{}                         & \multicolumn{1}{c|}{}                                                                                   & \multicolumn{1}{c|}{Gallery} & \multicolumn{1}{c}{Query} \\ \hline

Market-1501       & 751/12,936 & 750/19,732 & 750/3,368 \\
DukeMTMC-reID         & 702/16,522 & 1,110/17,661 & 702/2,228 \\ \hline
Occluded-Duke   & 702/15,618 & 1,110/17,661 & 519/2,210 \\ 
Occluded-ReID   & -      & 200/1,000  & 200/1,000 \\ \hline
Partial-REID    & -      & 60/300    & 60/300 \\
Partial-iLIDS   & -      & 119/119   & 119/119 \\ 
\hline
\hline
\end{tabular}
}
\end{center}
\caption{Dataset details. We extensively evaluate our proposed method on 6 public datasets, including 2 holistic, 2 occluded and 2 partial ones.}
\end{table}

\begin{table}[t]
\centering
\begin{tabular}{l|cc|cc}
\hline 
\hline 
\multicolumn{1}{l|}{\multirow{2}{*}{Methods}} & \multicolumn{2}{c|}{Occluded-Duke}            & \multicolumn{2}{c}{Occluded-REID}       \\ 
\multicolumn{1}{c|}{}                                 & \multicolumn{1}{c}{Rank-1} & \multicolumn{1}{c|}{mAP} & \multicolumn{1}{c}{Rank-1} & \multicolumn{1}{c}{mAP} 
\\ \hline 
 Part-Aligned~\cite{zhao2017deeply} & 28.8 & 20.2 & - & -\\
 PCB~\cite{sun2018beyond}  & 42.6 & 33.7 & 41.3 & 38.9 \\ 
 \hline
 Part Bilinear~\cite{suh2018part}  & 36.9 & - & - & - \\ 
 FD-GAN~\cite{ge2018fd}  & 40.8 & - & - & - \\ 
 \hline
 AMC+SWM~\cite{zheng2015partial}  & - & - & 31.2 & 27.3 \\
 DSR~\cite{he2018deep} & 40.8 & 30.4 & 72.8 & 62.8 \\
 SFR~\cite{he2018recognizing}  & 42.3 & 32 & - & - \\
 \hline
 Ad-Occluded~\cite{huang2018adversarially} & 44.5 & 32.2 & - & - \\
 TCSDO~\cite{zhuo2019novel} & - & - & 73.7 & 77.9 \\ 
 FPR~\cite{he2019foreground-aware}  & - & - & 78.3 & 68.0 \\
 PGFA~\cite{miao2019PGFA}  & 51.4 & 37.3 & - & - \\ 
 \hline
 \textbf{HOReID} (\textit{Ours})              &  \textbf{55.1} & \textbf{43.8} & \textbf{80.3} & \textbf{70.2} \\
 \hline     
 \hline   
\end{tabular}
\vspace{2pt}
\caption{Comparison with state-of-the-arts on two occluded datasets, \textit{i.e.} Occluded-Duke~\cite{miao2019PGFA} and Occluded-REID~\cite{zhuo2018occluded}.}
\label{tab:occluded_result}
\end{table}

\subsection{Experimental Results}

\textbf{Results on Occluded Datasets.}
We evaluate our proposed framework on two occluded datasets, \textit{i.e.} Occluded-Duke~\cite{miao2019PGFA} and Occluded-ReID~\cite{zhuo2018occluded}.
Occluded-Duke is selected from DukeMTMC-reID by leaving occluded images and filter out some overlap images.
It contains 15,618 training images, 17,661 gallery images, and 2,210 occluded query images.
Occluded-ReID is captured by the mobile camera, consist of 2000 images of 200 occluded persons. Each identity has five full-body person images and five occluded person images with different types of severe occlusions.

Four kinds of methods are compared, they are vanilla holistic ReID methods \cite{zhao2017deeply,sun2018beyond}, holistic ReID methods with key-points information \cite{suh2018part,ge2018fd}, partial ReID methods \cite{zheng2015partial,he2018deep,he2018recognizing} and occluded ReID methods \cite{huang2018adversarially,zhuo2019novel,he2019foreground-aware,miao2019PGFA}. 
The experimental results are shown in Table \ref{tab:occluded_result}.
As we can see, there is no significant gap between vanilla holistic ReID methods and holistic methods with key-points information. For example, PCB~\cite{sun2019perceive} and FD-GAN~\cite{ge2018fd} both achieve approximately $40\%$ Rank-1 score on Occluded-Duke dataset, showing that simply using key-points information may not significantly benefit occluded ReID task.
For partial ReID and occluded ReID methods, they both achieve an obvious improvement on occluded datasets. For example, DSR \cite{he2018deep} get a $72.8\%$ and FPR \cite{he2019foreground-aware} get a $78.3\%$ Rank-1 scores on Occluded-REID dataset. This shows that occluded and partial ReID task share similar difficulties, \textit{i.e.} learning discriminative feature and feature alignment.
Finally, our proposed framework achieves best performance on Occluded-Duke and Occlude-REID datasets at $55.1\%$ and $80.4\%$ in terms of Rank-1 score, showing the effectiveness.

\textbf{Results on Partial Datasets.}
Accompanied by occluded images, partial ones often occur due to imperfect detection, outliers of camera views, and so on.
To further evaluate our proposed framework, in Table \ref{tab:partial_result} we also report the results on two partial datasets, Partial-REID~\cite{zheng2015partial} and Partial-iLIDS~\cite{he2018deep}.
Partial-REID includes 600 images from 60 people, with five full-body images and five partial images per person, which is only used for the test. 
Partial-iLIDS is based on the iLIDS~\cite{he2018deep} dataset and contains a total of 238 images from 119 people captured by multiple non-overlapping cameras in the airport, and their occluded regions are manually cropped.
Following \cite{sun2019perceive,he2019foreground-aware,zhuo2019novel}, because the two partial datasets are too small, we use Market-1501 as training set and the two partial datasets as test set.
As we can see, our proposed framework significantly outperforms the other methods by at least $2.6\%$ and $4.4\%$ in terms of Rank-1 score on the two datasets.

\begin{table}[t]
\begin{center}
\scalebox{1}{
\begin{tabular}{l|cc|cc}
\hline
\hline 
\multicolumn{1}{l|}{\multirow{2}{*}{Methods}} & \multicolumn{2}{c|}{Partial-REID}            & \multicolumn{2}{c}{Partial-iLIDS}       \\ 
\multicolumn{1}{c|}{}                                 & \multicolumn{1}{c}{Rank-1} & \multicolumn{1}{c|}{Rank-3} & \multicolumn{1}{c}{Rank-1} & \multicolumn{1}{c}{Rank-3} \\ \hline
 DSR \cite{he2018deep}& 50.7 & 70.0 & 58.8 & 67.2 \\
 SFR \cite{he2018recognizing} & 56.9 & 78.5 & 63.9 & 74.8 \\
 VPM \cite{sun2019perceive} & 67.7 & 81.9 & 65.5 & 74.8 \\ 
 PGFA \cite{miao2019PGFA} & 68.0 & 80.0 & 69.1 & 80.9 \\
 AFPB \cite{zhuo2018occluded} & 78.5 & - & - & - \\ 
 FPR \cite{he2019foreground-aware} & 81.0 & - & 68.1 & - \\ 
 TCSDO \cite{zhuo2019novel}& 82.7 & - & - & - \\ 
 \hline
 \textbf{HOReID}(\textit{Ours})      & \textbf{85.3} & \textbf{91.0} & \textbf{72.6} & \textbf{86.4} \\
\hline
\hline
\end{tabular}
}
\end{center}
\caption{Comparison with state-of-the-arts on two partial datasets, \textit{i.e.} Partial-REID~\cite{zheng2015partial} and Partial-iLIDS~\cite{he2018deep} datasets. Our method achieves best performance on the two partial datasets.}
\label{tab:partial_result}
\end{table}

\begin{table}[t]
\begin{center}
\scalebox{1}{
\begin{tabular}{l|cc|cc}
\hline
\hline
\multicolumn{1}{l|}{\multirow{2}{*}{Methods}} & \multicolumn{2}{c|}{Market-1501} & \multicolumn{2}{c}{DukeMTMC} \\ 
\multicolumn{1}{c|}{} & \multicolumn{1}{c}{Rank-1} & \multicolumn{1}{c|}{mAP} & \multicolumn{1}{c}{Rank-1} & \multicolumn{1}{c}{mAP} \\ 
\hline 
 PCB \cite{sun2018beyond} & 92.3 & 77.4 & 81.8 & 66.1 \\
 VPM \cite{sun2019perceive} & 93.0 & 80.8 & 83.6 & 72.6 \\  
 BOT \cite{luo2019bag} & 94.1 & 85.7 & 86.4 & 76.4 \\
 \hline
 SPReID \cite{kalayeh2018human}& 92.5 & 81.3 & - & - \\
 MGCAM \cite{song2018mask} & 83.8 & 74.3 & 46.7 & 46.0 \\
 MaskReID \cite{qi2018maskreid} & 90.0 & 75.3 & - & -\\
 FPR \cite{he2019foreground-aware} & 95.4 & 86.6 & 88.6 & 78.4 \\ 
 \hline
 PDC \cite{su2017pose} &  84.2 & 63.4 & - & - \\
 Pose-transfer \cite{liu2018pose} & 87.7 & 68.9 & 30.1 & 28.2 \\
 PSE \cite{saquib2018pose} & 87.7 & 69.0 & 27.3 & 30.2 \\
 PGFA \cite{miao2019PGFA} & 91.2 & 76.8 & 82.6 & 65.5 \\ 
 \hline
 \textbf{HOReID}(\textit{Ours}) & 94.2 & 84.9 & 86.9 & 75.6 \\
 \hline
 \hline
\end{tabular}
}
\end{center}
\caption{Comparison with state-of-the-arts on two holistic datasets, Market-1501 \cite{zheng2015scalable} and DukeMTMTc-reID \cite{ristani2016performance,zheng2017unlabeled}. Our method achieves comparable performance on holistic ReID.}
\label{tab:holistic_results}
\end{table}

\textbf{Results on Holistic Datasets.}
Although recent occluded/partial ReID methods have obtained improvements on occluded/partial datasets, they often fails to get a satisfying performance on holistic datasets. This is caused by the noise during feature learning and alignment.
In this part, we show that our proposed framework can also achieve satisfying performance on holistic ReID datasets including Market-1501 and DuekMTMTC-reID.
Market-1501~\cite{zheng2015scalable} contains 1,501 identities observed from 6 camera viewpoints, 19,732 gallery images and 12,936 training images, all the dataset contains few of occluded or partial person images.
DukeMTMC-reID~\cite{ristani2016performance,zheng2017unlabeled} contains 1,404 identities, 16,522 training images, 2,228 queries, and 17,661 gallery images.

Specifically, we conduct experiments on two common holistic ReID datasets Market-1501 \cite{zheng2015scalable} and DukeMTMC-reID \cite{ristani2016performance,zheng2017unlabeled}, and compare with 3 vanilla ReID methods \cite{sun2018beyond,sun2019perceive,luo2019bag}, 3 ReID methods with human-parsing information \cite{kalayeh2018human,song2018mask,qi2018maskreid,he2019foreground-aware} and 4 holistic ReID methods with key-points information \cite{su2017pose,liu2018pose,saquib2018pose,miao2019PGFA}.
The experimental results are shown in Table \ref{tab:holistic_results}. 
As we can see, the 3 vanilla holistic ReID methods obtain very competitive performance. For example, BOT \cite{luo2019bag} gets a $94.1\%$ and $86.4\%$ Rank-1 score on two datasets.
However, for the holistic ReID methods using external cues such human-parsing and key-points information perform worse. For example, SPReID \cite{kalayeh2018human} uses human-parsing information and only achieves $92.5\%$ Rannk-1 score on Market-1501 dataset. PFGA \cite{miao2019PGFA} uses key-points information and only gets a $82.6\%$ Rank-1 score on DukeMTMC-reID dataset.
This shows that simply using external cues such as human-parsing and key-points may not bring improvement on holistic ReID datasets. 
This is caused by that the most images holistic ReID datasets are well detected, vanilla holistic ReID methods is powerful enough to learn discrimintive features.
Finally, we propose a adaptive direction graph convolutional (ADGC) layer which can suppress noisy features and a cross-graph embedded-alignment (CGEA) layer which can avoid hard one-to-one alignment. With the proposed ADGC and CGEA layers, our framework also achieves comparable performance on the two holistic ReID datasets. Specifically, we achieve about $94\%$ and $87\%$ Rank-1 scores on Market-1501 and DukeMTMC-reID datasets.

\subsection{Model Analysis}

\begin{table}[]
\begin{center}
\scalebox{1}{
\centering
\begin{tabular}{c|ccc|c|c}
\hline
\hline
    Index & $\mathcal{S}$ & $\mathcal{R}$ & $\mathcal{T}$ & Rank-1 & mAP \\ \hline
    1 & $\times$ & $\times$ & $\times$ & 49.9 & 39.5 \\
    2 & $\checkmark$ & $\times$ & $\times$ & 52.4 & 42.8 \\ 
    3 & $\checkmark$ & $\checkmark$ & $\times$ & 53.9 & 43.2 \\ 
    4 & $\checkmark$ & $\checkmark$ & $\checkmark$ & \textbf{55.1} & \textbf{43.8} \\
\hline
\hline
\end{tabular}
}
\end{center}
\caption{
Analysis of one-order semantic module ($\mathcal{S}$), high-order relation module ($\mathcal{R}$) and high-order human-topology module ($\mathcal{T}$). The experimental results show the effectiveness of our proposed three modules.}
\label{tab:module-analysis}
\end{table}

\textbf{Analysis of Proposed Modules.} 
In this part, we analyze our proposed one-order semantic module ($\mathcal{S}$), high-order relation module ($\mathcal{R}$) and high-order human-topology module ($\mathcal{T}$). The experimental results are shown in Table \ref{tab:module-analysis}.
Firstly, in index-1, we remove all the three modules degrading our framework to an IDE model \cite{zheng2016person}, where only a global feature $V_g$ is available. Its performance is unsatisfying and only achieves $49.9\%$ Rank-1 score.
Secondly, in index-2, when using one-order semantic information, the performance is improved by $2.5\%$ and up to $52.4\%$ Rank-1 score. This shows that the semantic information from key-points is useful for learning and aligning features.
Thirdly, in index-3, extra high-order relation information is added, and the performance is further improved by $1.5\%$ achieving $53.9\%$. This demonstrates the effectiveness of our module $\mathcal{R}$.
Finally, in index-4, our full framework achieves the best accuracy at $55.1\%$ Rank-1 score, showing the the effectiveness of our module $\mathcal{T}$.

\textbf{Analysis of Proposed layers.}
In this part, we further analyze normalization of key-point confidences (NORM), adaptive direction graph convolutional (ADGC) layer and cross-graph embedded-alignment (CGEA) layer, which are the key components of for semantic module ($\mathcal{S}$), relation module ($\mathcal{R}$) and topology module ($\mathcal{T}$).
Specifically, when removing NORM, straightly use the original confidence score. 
When removing ADGC, in Eq.(\ref{eq:adgc}), we replace $A^{adj}$ with a fixed adjacency matrix linked like a human-topology. Thus, the relation module ($\mathcal{S}$) degrades to a vanilla GCN, which cannot suppress noise information. 
When removing CGEA, in Eq.(\ref{eq:CGEA-layer}), we replace $U_1$ and $U_2$ with a fully-connected matrix. That is, every node of graph 1 is connected to all nodes of graph 2. Then, the topology module ($\mathcal{T}$) contains no high-order human-topology information for feature alignment and degrades to a vanilla verification module.
The experimental results are shown in Table \ref{tab:components-analysis}.
As we can see, when removing NORM, ADGC or CGEA, the performance significantly drop by $2.6\%$, $1.4$\% and $0.7\%$ rank-1 scores. The experimental results show the effectiveness of our proposed NORM, ADGC and CGEA components.

\begin{table}[]
\begin{center}
\scalebox{1}{
\centering
\begin{tabular}{ccc|c|c}
\hline \hline
    NORM & ADGC & CGEA & Rank-1 & mAP \\ \hline
    $\times$ & $\checkmark$ & $\checkmark$   & 52.5 & 40.4 \\ 
    $\checkmark$ & $\times$ & $\checkmark$   & 53.7 & 42.2 \\
    $\checkmark$ & $\checkmark$ & $\times$   & 54.4 & 43.5 \\
    $\checkmark$ & $\checkmark$ & $\checkmark$   & \textbf{55.1} & \textbf{43.8} \\
\hline \hline
\end{tabular}
}
\end{center}
\caption{Analysis of normalization of key-point confidences (NORM), adaptive direction graph convolutional (ADGC) layer and cross-grpah embedded-alignment (CGEA) layer. The experimental results show the effectiveness of our proposed layers.}
\label{tab:components-analysis}
\end{table}

\begin{figure}[t]
\centering
\includegraphics[scale=0.32]{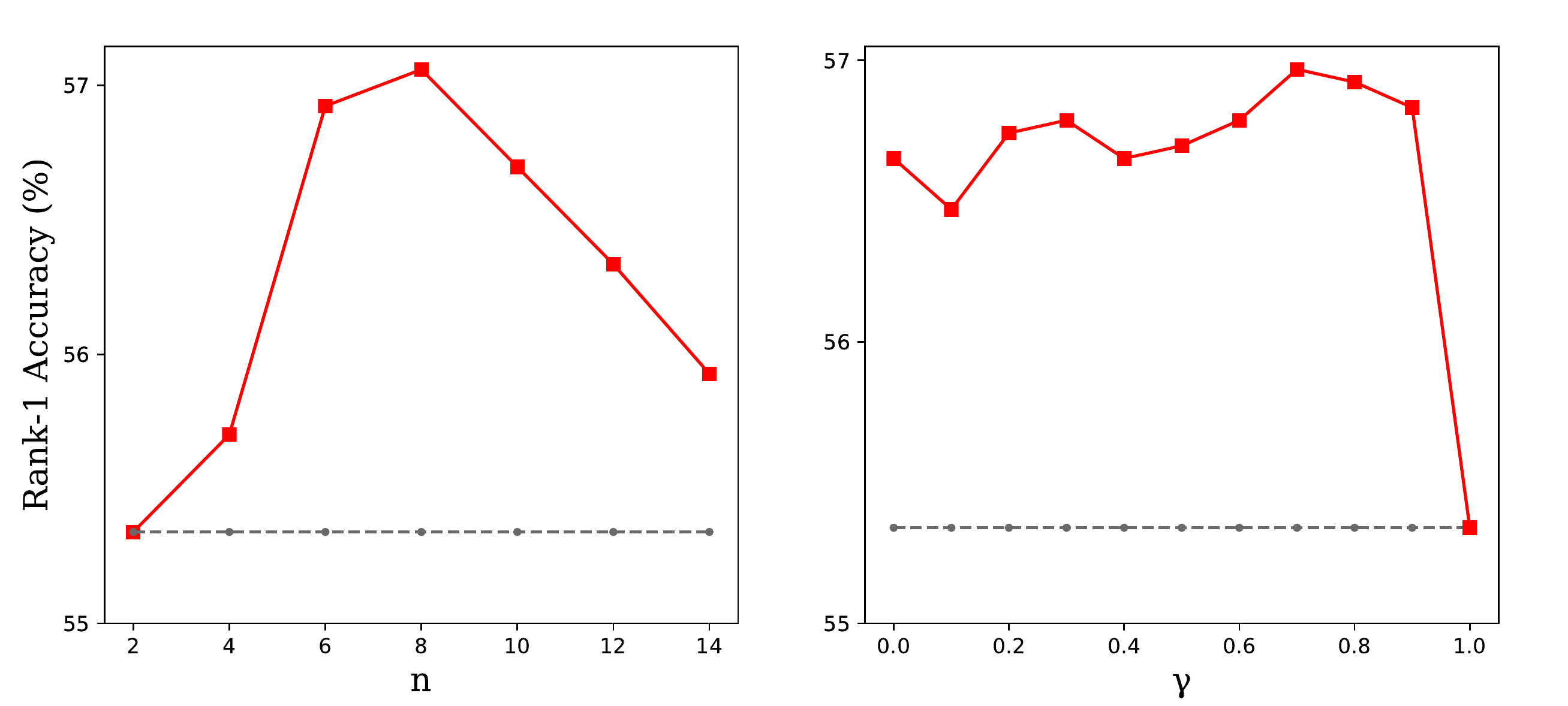}
\caption{Analysis of parameters $\gamma$ and $n$ in Eq.(\ref{eq:final-similarity}). The optimal values are $\gamma=0.5$ and $n=8$. When analyze one of them, the other one is fixed as its optimal value. The The experimental results shows that our model is robust to different parameters.}
\label{fig:param}
\end{figure}

\textbf{Analysis of Parameters.}
We evaluate the effects of parameters in Eq.(\ref{eq:final-similarity}), \textit{i.e.} $\gamma$ and $n$. 
The results are shown Figure \ref{fig:param}, and the optimal setting is $\gamma=0.5$ and $n=8$. 
When analyzing one parameter, the other is fixed at the optimal value. 
It is clear that, when using different $\gamma$ and $n$, our model stably outperforms the baseline model.
The experimental results show the our proposed framework is robust to different weights. Please note that the performances here are different from Table \ref{tab:occluded_result}, where the former achieves $57\%$ while and the latter $55\%$. This is because the latter is computed using average of 10 times for fair comparison. 

\section{Conclusion}
In this paper, we propose a novel framework to learn high-order relation information for discriminative features and topology information for robust alignment.
For learning relation information, we formulate local features of an image as nodes of a graph and propose an adaptive-direction graph convolutional (ADGC) layer to promote the message passing of semantic features and suppress that of meaningless and noisy ones.
For learning topology information, we propose a cross-graph embedded-alignment (CGEA) layer conjugated with a verification loss, which can avoid sensitive hard one-to-one alignment and perform a robust soft alignment.
Finally, extensive experiments on occluded, partial and holistic datasets demonstrate the effectiveness of our proposed framework.

\section*{Acknowledge}
This research was supported by National Key R\&D Program of China (No. 2017YFA0700800).

\normalem
{\small
\bibliographystyle{ieee_fullname}
\bibliography{egbib}
}

\end{document}